\documentclass[sigconf]{acmart}

\usepackage{booktabs} 
\usepackage{amsmath,amssymb,multicol, mathrsfs}
\usepackage[ruled,linesnumbered]{algorithm2e}
\usepackage{graphicx}
\usepackage{balance}
\usepackage{epstopdf}

\usepackage{multirow}
\usepackage{url}
\usepackage{subfigure}
\usepackage[flushleft]{threeparttable}
\usepackage[utf8]{inputenc}
\usepackage{booktabs,tabularx}

\usepackage{multirow}
\usepackage{color,soul}

\renewcommand\vec[1]{\overrightarrow{#1}}
\newcommand\cev[1]{\overleftarrow{#1}}

\usepackage{enumitem}
\setlist{leftmargin=3mm}

\AtBeginDocument{%
  \providecommand\BibTeX{{%
    \normalfont B\kern-0.5em{\scshape i\kern-0.25em b}\kern-0.8em\TeX}}}

\setcopyright{acmcopyright}
\copyrightyear{2020}
\acmYear{2020}
\setcopyright{iw3c2w3}
\acmConference[WWW '20]{Proceedings of The Web Conference 2020}{April 20--24, 2020}{Taipei, Taiwan}
\acmBooktitle{Proceedings of The Web Conference 2020 (WWW '20), April 20--24, 2020, Taipei, Taiwan}
\acmPrice{}
\acmDOI{10.1145/3366423.3380120}
\acmISBN{978-1-4503-7023-3/20/04}
\begin{document}

\title[Text-to-SQL Generation for Question Answering on Electronic Medical Records]{Text-to-SQL Generation for Question Answering \\on Electronic Medical Records}


\author{Ping Wang}
\affiliation{%
 \institution{Dept. of Computer Science}
 \city{Virginia Tech, Arlington, VA}}
\email{ping@vt.edu}

\author{Tian Shi}
\affiliation{
  \institution{Dept. of Computer Science}
  \city{Virginia Tech, Arlington, VA}
}
\email{tshi@vt.edu}

\author{Chandan K. Reddy}
\affiliation{
  \institution{Dept. of Computer Science}
  \city{Virginia Tech, Arlington, VA}
}
\email{reddy@cs.vt.edu}

\renewcommand{\shortauthors}{P. Wang, et al.}

\begin{abstract}
Electronic medical records (EMR) contain comprehensive patient information and are typically stored in a relational database with multiple tables. 
Effective and efficient patient information retrieval from EMR data is a challenging task for medical experts.
\textit{Question-to-SQL generation} methods tackle this problem by first predicting the SQL query for a given question about a database, and then, executing the query on the database.
However, most of the existing approaches have not been adapted to the healthcare domain due to a lack of healthcare Question-to-SQL dataset for learning models specific to this domain.
In addition, wide use of the abbreviation of terminologies and possible typos in questions introduce additional challenges for accurately generating the corresponding SQL queries.
In this paper, we tackle these challenges by developing a deep learning based \textbf{TR}anslate-\textbf{E}dit Model for \textbf{Q}uestion-to-\textbf{S}QL~(TREQS) generation, which adapts the widely used sequence-to-sequence model to directly generate the SQL query for a given question, and further performs the required edits using an attentive-copying mechanism and task-specific look-up tables. 
Based on the widely used publicly available electronic medical database, we create a new large-scale Question-SQL pair dataset, named \textit{MIMICSQL}, in order to perform the Question-to-SQL generation task in healthcare domain. 
An extensive set of experiments are conducted to evaluate the performance of our proposed model on MIMICSQL. Both quantitative and qualitative experimental results indicate the flexibility and efficiency of our proposed method in predicting condition values and its robustness to random questions with abbreviations and typos. 
\end{abstract}

\ccsdesc[500]{Information systems~Question answering}
\ccsdesc[500]{Computing methodologies~Information extraction}
\ccsdesc[500]{Computing methodologies~Neural networks}
\ccsdesc[500]{Applied computing~Health informatics}
\vspace{0.1in}
\keywords{Sequence-to-sequence model, attention mechanism, pointer-generator network, electronic medical records, SQL query.}


\maketitle
\section{Introduction}
\label{sec1}

Due to recent advances of data collection and storing techniques, a large amount of healthcare related data, typically in the form of electronic medical records (EMR), are accumulated everyday in clinics and hospitals.
EMR data contain a comprehensive set of longitudinal information about patients and are usually stored in structured databases with multiple relational tables, such as demographics, diagnosis, procedures, prescriptions, and laboratory tests. 
One important mechanism of assisting doctors' clinical decision making is to directly retrieve patient information from EMR data, including patient-specific information (e.g., individual demographic and diagnosis information) and cohort-based statistics (e.g., mortality rate and prevalence rate).
Typically, doctors interact with EMR data using searching and filtering functions available in rule-based systems that first turn any predefined-rule (front-end) to a SQL query (back-end), and then, return an answer.
These systems are complicated and require special training before being used.
They are also difficult to manage and extend. 
For example, the front-end needs to be adapted for newer functionalities.
Therefore, doctors who depend on these systems cannot fully and freely explore EMR data.
Another challenge for these systems is that the users have to first transform their questions to a combination of rules in the front-end, which is not convenient and efficient.
For instance, if a doctor wants to know the number of patients who are under the age of 40 and suffering from diabetes, then, he/she may have to create two filters, one for disease and the other for age. An alternate way to solve this problem is to build a model that can translate this question directly to its SQL query, so that the doctor only needs to type his/her question as: ``Give me the number of patients whose age is below 40 and have diabetes", in the search box to get the answer. 
Motivated by this intuition, we propose a new deep learning based model that can translate textual questions on EMR data to SQL queries (\textit{Text-to-SQL generation}) without any assistance from a database expert.
As a result, these systems can assist doctors with their clinical decisions more efficiently. Since the textual input in our task is a clinical question, we will refer to Text-to-SQL generation as \textit{Question-to-SQL generation} from now onwards.

Recently, Question-to-SQL generation task has gained significant attention and found applications in a variety of domains,
including WikiSQL~\cite{dong2018coarse,zhong2017seq2sql,xu2017sqlnet} for Wikipedia, ATIS~\cite{iyer2017learning} about flight-booking, GeoQuery~\cite{finegan2018improving} about US geography, and Spider~\cite{yu2018spider} for general purpose cross-domain applications.
There are also few works in this line of research in healthcare domain~\cite{ben2012medical,roberts2017semantic}.
Broadly speaking, various approaches for Question-to-SQL generation task belong to one of the following two categories:
(1) \textit{Semantic parsing or slot-filling methods}~\cite{xu2017sqlnet,yu2018typesql,dong2018coarse,yavuz2018takes,Guo2019TowardsCT,yu2018syntaxsqlnet}: These models make use of semantic and syntactic information in a question and a table schema to generate a SQL logic form (parsing tree), which can be easily converted to the corresponding executable query. However, they strongly depend on pre-defined templates, which limits their applications in generating complex SQL queries.
(2)~\textit{Language generation methods} \cite{zhong2017seq2sql,wang2018robust,zhang2019editing}: These models leverage the power of language generation models and can directly generate SQL queries 
without building pre-defined SQL templates~\cite{mccann2018natural}.
Therefore, they can be easily applied to produce complex SQL queries, regardless of the number of tables and columns involved.
However, the predicted SQL queries may not be executable due to limited and inaccurate information from questions (e.g., a random question with typos and missing keywords).
Moreover, it is difficult to interpret language generation models and their outputs.

\begin{figure}[!tp]
    \vspace{-2mm}
	\centering
	\includegraphics[width=0.4\textwidth,height=65mm]{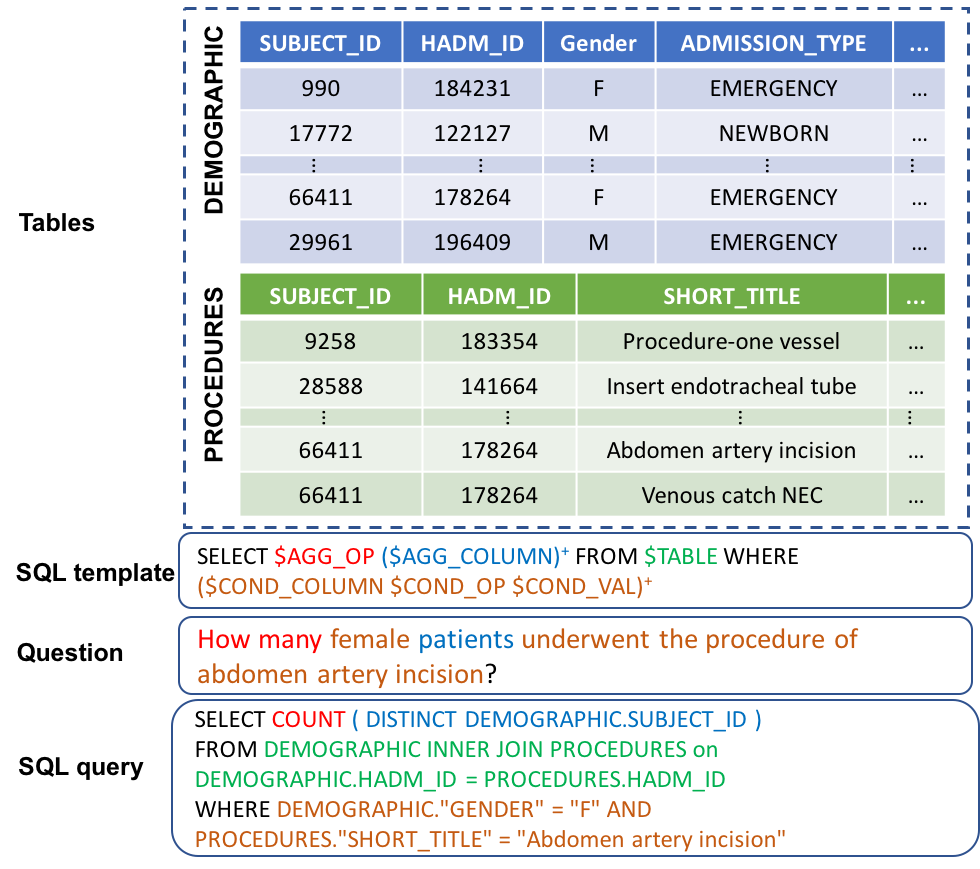}
	\vspace{-2.5mm}
	\caption{An example from MIMICSQL. The two tables, namely, Demographics and Diagnoses, are used to answer the question. Different colors are used to show the correspondence between various components in source question, targeted SQL query, and SQL template.}
	\vspace{-8mm}
	\label{fig:sample}
\end{figure}

Many recent Question-to-SQL models have been primarily benchmarked on WikiSQL~\cite{dong2018coarse,zhong2017seq2sql,xu2017sqlnet} and Spider~\cite{Guo2019TowardsCT,yu2018syntaxsqlnet,bogin2019representing} datasets.
In WikiSQL, questions are asked against databases with a single table, and every SQL query is composed of a ``SELECT'' (column/aggregation) clause along with a ``WHERE'' (condition) clause that consists of one or multiple conditions.
Different from WikiSQL, the Spider dataset contains lots of complex and nested queries (e.g., ``GROUP BY'' and ``HAVING'') which may involve multiple tables \cite{yu2018spider}.
Some recent studies have shown that several models which perform well on WikiSQL achieve poor results on Spider \cite{zhang2019editing,Guo2019TowardsCT}.
It indicates that models for Question-to-SQL generation on single-table databases cannot be simply adapted to database with multiple relational tables.
For Spider dataset, the current Question-to-SQL generation task focuses on generating SQL queries without actual values for ``WHERE'' conditions,
which means models are only required to predict SQL structures and parse corresponding table and column names.
However, even if a model can produce high quality SQL structures and columns, condition value generation may still be the bottleneck in producing correct and executable SQL queries~\cite{yavuz2018takes}.
Another issue with WikiSQL and Spider dataset is that most words ($78\%$ for WikiSQL and $65\%$ for Spider) in database schema in development/testing sets have appeared in the training set \cite{Guo2019TowardsCT}.
Therefore, it is not feasible to apply the models trained on the Spider dataset to some other domains like chemistry, biology, and healthcare.
Specific to healthcare domain, 
Question-to-SQL generation for EMR data is still under-explored. 
There are three primary challenges:
(1) \textbf{Medical terminology abbreviations.} Due to the wide use of abbreviation of medical terminology (sometimes typos), it is difficult to match keywords in questions to those in database schema and table content.
(2) \textbf{Condition value parsing and recovery.} 
It is still a challenging task to extract condition values from questions and recover them based on table content, especially in the appearance of medical abbreviations.
(3) \textbf{Lack of large-scale healthcare Question-to-SQL dataset.} Currently, there is no dataset available for the Question-to-SQL task in the healthcare domain.

To tackle these challenges, we first generated a large-scale healthcare Question-to-SQL dataset, namely MIMICSQL, that consists of $10,000$ Question-SQL pairs, by using the publicly available real-world Medical Information Mart for Intensive Care III (MIMIC III) dataset~\cite{johnson2016mimic,goldberger2000physiobank} and leveraging the power of \textit{crowd-sourcing}. An illustrative example in MIMICSQL is provided in Figure~\ref{fig:sample} to illustrate various components of the dataset. 
Based on MIMICSQL data, we further propose a language generation based Translate-Edit model,
which can first translate a question to the corresponding SQL query, and then, retrieve condition values based on the question and table content.
The editing meta-algorithms make our model more robust to randomly asked questions with insufficient information and typos, and make it practical to retrieve and recover condition values effectively. 
The major contributions of this paper are as follows:

\begin{itemize}[leftmargin=*]
\item Propose a two-stage \textbf{TR}anslate-\textbf{E}dit Model for \textbf{Q}uestion-to-\textbf{S}QL (TREQS) generation model, which consists of three main components: (1) Translating an input question to a SQL query using a Seq2Seq based model, (2) Editing the generated query with attentive-copying mechanism, and (3) Further editing it with task-specific look-up tables.

\item Create a large-scale dataset for Question-to-SQL task in healthcare domain. 
MIMICSQL has two subsets, in which the first set is composed of template questions (machine generated), while the second consists of natural language questions (human annotated).
To the best of our knowledge, it is the first dataset for healthcare question answering on EMR data with multi-relational tables.

\item 
Conduct an extensive set of experiments on MIMICSQL dataset for both template questions and natural language questions to demonstrate the effectiveness of the proposed model. 
Both qualitative and quantitative results indicate that it outperforms several baseline methods. 

\end{itemize}
The rest of this paper is organized as follows. Section~\ref{sec2} describes some prior work related to Question-to-SQL generation, and differentiate our work from other existing works. Section~\ref{sec3} provides a comprehensive description of the MIMICSQL data generation process. Section~\ref{sec4} provides the details of the proposed translate-edit model. Section~\ref{sec5} shows the comparison of our proposed model with the state-of-the-art methods by analyzing both quantitative and qualitative results. 
Finally, we conclude the paper in Section~\ref{sec6}.
\vspace{-2mm}
\section{Related Work}
\label{sec2}
Question-to-SQL generation is a sub-task of semantic parsing, which aims at translating a natural language text to a corresponding formal semantic representation, including SQL queries, logic forms and code generation~\cite{yaghmazadeh2017sqlizer, dong2016language}.
It has attracted significant attention in various applications,
including WikiSQL~\cite{zhong2017seq2sql,xu2017sqlnet} for Wikipedia, ATIS~\cite{iyer2017learning} about flight-booking, GeoQuery~\cite{finegan2018improving} about US geography and Spider~\cite{yu2018spider} about cross-domain.
In the literature of Question-to-SQL generation, a common way is to utilize a SQL structure-based sketch with multiple slots and formulate the problem as a slot filling task~\cite{dong2018coarse,zhong2017seq2sql,xu2017sqlnet,yu2018typesql,shi2018incsql} by incorporating some form of pointing/copying mechanism \cite{vinyals2015pointer}. 
Seq2SQL method~\cite{zhong2017seq2sql} is an augmented pointer network based framework and mainly prunes the output space of the target query by leveraging the unique structures of SQL commands. 
SQLNet method~\cite{xu2017sqlnet} is proposed to avoid the ``order-matter'' problem in the condition part by using a sketch-based approach instead of the sequence-to-sequence (Seq2Seq) based method. 
By further improving SQLNet, TYPESQL method~\cite{yu2018typesql} captures the rare entities and numbers in natural language questions by utilizing the type information. 
The two-stage semantic parsing method named Coarse2Fine~\cite{dong2018coarse} first generates a sketch of a given question and then fills in missing details based on both input question and the sketch. 
Recently, several semantic parsing methods~\cite{Guo2019TowardsCT,yu2018syntaxsqlnet,bogin2019representing} are also proposed on Spider to tackle the problem across different domains.
One limitation of these methods is that they are highly dependent on the SQL structure and the lexicons, and thus cannot efficiently retrieve the condition values. Therefore, compared to other components, the performance of most semantic parsing methods in predicting condition values tend to be relatively low and these methods primarily focus on predicting correct SQL structures and columns, especially for the cross-domain problem present in the recently released  Spider data~\cite{yu2018spider}.

To overcome the disadvantage of slot filling methods,  Seq2Seq based methods~\cite{sutskever2014sequence,dong2016language,wang2018pointing,mccann2018natural} are proposed to tackle this challenge by directly generating the targeted SQL queries. More specifically, Seq2Seq based methods first encode input questions into vector representations and then decode the corresponding SQL conditioned on the encoded vectors. A type system of SQL expressions is applied in the deep Seq2Seq model in~\cite{wang2018pointing} to guide the decoder to either directly generate a token from the vocabulary or copy it from the input question. The table schema and the input question are encoded and concatenated as the model input. In contrast, the column names are encoded independently from the encoding of questions in~\cite{lukovnikov2018translating}, which extended the pointer-generator in SQL generation when the order of conditions in SQL query does not matter. In~\cite{mccann2018natural}, a unified question-answering framework was proposed to handle ten different natural language processing tasks, including WikiSQL semantic parsing task.
To perform question answering on databases with multiple relational tables, there are some other works that aim at guiding the SQL generation indirectly using the answers obtained by query execution~\cite{yin2015neural,pasupat2015compositional,yih2015semantic} or accomplish the goal by directly identifying the correct table cells corresponding to the question answers~\cite{sun2016table, guo2019table2answer}. 

Both semantic parsing and language generation approaches show great efficiency in the existing application domains. However, the Question-to-SQL generation task in healthcare domain is still under-explored. There are some efforts in directly seeking answers from unstructured clinical notes to assist doctors with their clinical decision making~\cite{yu2007development,lee2006beyond}. However, these problems are significantly different from our task of answering natural language questions on structured EMR data since in our task, the answers to the questions may not be directly included in the structured data. For example, instead of directly retrieving answers, a certain extent of reasoning is required to answer the counting questions starting with ``\textit{how many}".    There are a few research efforts in solving the Question-to-SQL generation tasks in healthcare domain using semantic parsing and named entity extraction ~\cite{ben2012medical,roberts2017semantic}. 
Due to the domain-specific challenges and the lack of large-scale datasets for model training, there are still several challenges for the Question-to-SQL generation in healthcare. For example, due to the commonly occurring abbreviations of healthcare terminology in EMR data and potential typos in questions, it is possible that the keywords provided in questions are not exactly the same ones used in the EMR data. Therefore, besides predicting the SQL structure and columns, one important task in healthcare is correctly predicting condition values in order to ensure the accuracy of query results for input questions.
These challenges motivate us to develop a model that can tackle these issues specifically in healthcare.
To train and test our model, we also create the MIMICSQL dataset, which consists of Question-SQL pairs based on MIMIC III dataset. This is the first work that focuses on the Question-to-SQL generation on the healthcare databases with multiple relational tables.
\vspace{-2mm}
\section{MIMICSQL Dataset Creation}
\label{sec3}
To the best of our knowledge, there is no existing dataset for Question-to-SQL generation task in the healthcare domain. In this section, we provide a detailed illustration of Question-SQL pair generation for Question-to-SQL tasks on EMR data. 

\vspace{-3mm}
\subsection{MIMIC III Dataset}
To ensure both the public availability of the dataset and the reproducibility of the results for Question-to-SQL generation methods, the widely used Medical Information Mart for Intensive Care III~(MIMIC~III) dataset~\cite{johnson2016mimic,goldberger2000physiobank} is used in this paper to create the Question-SQL pairs. Typically, the healthcare related patient information is grouped into five categories in healthcare literature, including demographics (Demo), laboratory tests (Lab), diagnosis (Diag), procedures (Pro), and prescriptions (Pres). We extracted patient information and prepared a specific table for each category separately. 
These tables compose a relational patient database where tables are linked through patient ID and admission ID as shown on the top of Figure~\ref{fig:sample}. 

\vspace{-3mm}
\subsection{MIMICSQL Generation}
Based on the aforementioned five tables, we create the MIMICSQL dataset, including the Question-SQL pairs along with the logical format for slot filling methods, specifically for such Question-to-SQL generation task. Figure~\ref{fig:sample} provides an overview of basic components used for MIMICSQL generation. Due to the large amount of information included in EMR database, it is challenging and time-consuming for domain experts to manually generate the Question-SQL pairs. 
It should be noted that, for machine generated questions, there exists some drawbacks, including not being natural compared to questions provided by humans and usually are not grammatically accurate. In this paper, we take advantage of both human and machine generation to collect the Question-SQL pairs for the MIMICSQL dataset in the following two steps.

\begin{figure*}[!tp]
\vspace{-4mm}
\subfigure[Dist. of length of template questions.]{\includegraphics[width=0.235\textwidth]{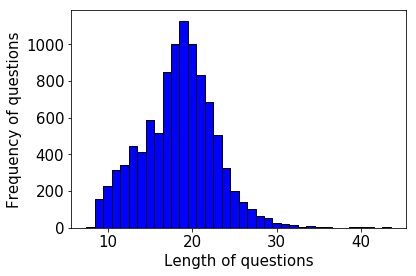}}
\subfigure[Dist. of length of NL questions.]{\includegraphics[width=0.235\textwidth]{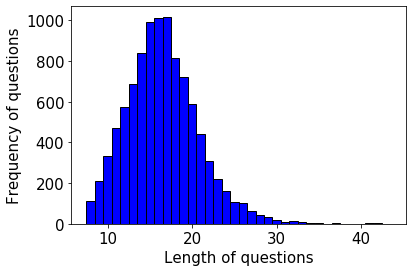}}
\subfigure[Dist. of length of SQL queries.]{\includegraphics[width=0.235\textwidth]{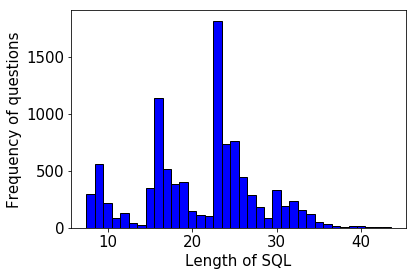}}
\subfigure[Dist. of No. of Questions.]{\includegraphics[width=0.235\textwidth]{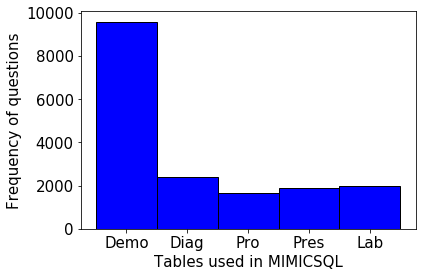}}
\vspace{-3mm}
\caption{Distribution of questions and queries in MIMICSQL dataset. ``Dist." is used as an acronym for ``Distribution". }
\vspace{-2mm}
\label{fig:stat2}
\end{figure*}

\subsubsection{Machine Generation of Questions}
Following the question types used in~\cite{kafle2018dvqa}, there are two types of questions in MIMICSQL, including retrieval questions and reasoning questions.
Following the generation of question templates in~\cite{pampari2018emrqa}, we first identify the questions that are possibly asked on the EMR data and then normalize them by identifying and replacing the entities regarding table headers, operations, and condition values with generic placeholders. The question templates for retrieval and reasoning questions are finally integrated into two generic templates. These question templates provide a guidance regarding the question topics or perspectives for the machine generated questions. 

\begin{enumerate}[leftmargin=*]
\vspace{-1mm}
    \item \textbf{Retrieval questions} are designed to directly retrieve specific patient information from tables. The two generic templates mainly used for retrieval questions include:
    \begin{itemize}[leftmargin=*]
        \item What is the $H_1$ and $H_2$ of Patient $Pat$ (or Disease $D$, or Procedure $Pro$, or Prescription $Pre$, or Lab test $L$)?
        \item List all the Patients (or Disease, or Procedures, or medications, or lab tests) whose $H_1\ O_1\ V_1$ and $H_2\ O_2\ V_2$.
    \end{itemize}
    
    \item \textbf{Reasoning questions} are designed to indirectly collect patient information by combining different components of five tables. The templates mainly used for reasoning questions include:
    \begin{itemize}[leftmargin=*]
        \item How many patients whose $H_1\ O_1\ V_1$ and $H_2\ O_2\ V_2$? 
        \item What is the maximum (or minimum, or average) $H_1$ of patient whose $H_2\ O_2\ V_2$ and $H_3\ O_3\ V_3$?  
    \end{itemize}
\end{enumerate}

Here, $H_i, O_i, V_i$ represent placeholders for the $i^{th}$ table column used in the question, its corresponding operation and condition value, respectively. In order to avoid complicated query structure, the number of conditions in each question cannot exceed a pre-defined threshold, which is set to be $2$ in this work.

During question generation, the corresponding SQL query for each question is also generated simultaneously.  
In order to respond to all questions without changing the query structure and facilitate the prediction of SQL for Question-to-SQL models, we adopt a general \textbf{SQL template} SELECT \$AGG\_OP (\$AGG\_COLUMN)$^+$ FROM \$TABLE WHERE (\$COND\_COLUMN \$COND\_OP \$COND\_VAL)$^+$.
Here, the superscript ``+'' indicates that it allows one or more items.
\textit{AGG\_OP} is the operation used for the selected AGG\_COLUMN and takes one of the five values, including ``NULL'' (representing no aggregation operation), ``COUNT'', ``MAX'', ``MIN'' and ``AVG''. 
\textit{AGG\_COLUMN} is the question topic that we are interested in each question and is stored as the column header in tables.
Since it is possible for a given question to be related to more than one table, \textit{TABLE} used here can be either a single table or a new table obtained by joining different tables. 
The part after WHERE represents the various conditions present in the question and each condition takes the form of (\$COND\_COLUMN \$COND\_OP \$COND\_VAL).
During query generation, we mainly consider five different condition operations, including ``$=$'', ``$>$'', ``$<$'', ``$>=$'' and ``$<=$''.

\begin{table}[!htp]
\vspace{-2mm}
\centering
\caption{Statistics of MIMICSQL dataset.}
\begin{threeparttable}
\vspace{-2mm}
\renewcommand\arraystretch{1.0}
\begin{tabular}{l|c}\hline
\textbf{Data}& \textbf{Value}  \\\hline
\# of patients &46,520  \\
\# of tables & 5 \\
\# of columns in tables\tnote{a} & 23/5/5/7/9 \\
\# of Question-SQL pairs& 10,000 \\
Average template question length (in words) &18.39 \\
Average NL question length (in words) &16.45\\
Average SQL query length & 21.14\\
Average aggregation columns &1.10 \\
Average conditions &1.76 \\\hline 
\end{tabular}
\begin{tablenotes}
\item[a] \footnotesize{The tables are in the order of Demographics, Diagnosis, Procedure, Prescriptions, and Laboratory tests.}
\end{tablenotes}
\end{threeparttable}
\vspace{-6mm}
\label{tab:stat} 
\end{table}

\subsubsection{Natural Language Question Collection}
These machine generation criteria make it practical to effectively obtain a set of Question-SQL pairs, however, there are two main drawbacks for the machine generated template questions. On the one hand, the questions may not be realistic in the clinical practice. For example, the unreasonable question ``\textit{How many patients whose primary disease is newborn and marital status is married?}" will also be generated.
On the other hand, the template questions tend to be not as natural as questions asked by doctors since they follow a fixed structure provided in the question templates. In order to overcome these drawbacks, we recruited eight Freelancers with medical domain knowledge on a \textit{\textit{crowd-sourcing platform named Freelancer}\footnote{www.freelancer.com}} to filter and paraphrase the template questions in three steps: 
(1) To ensure that the generated questions are realistic in the healthcare domain, each machine generated question is validated to ignore the unreasonable template questions.
(2) Each selected template question is rephrased as its corresponding natural language (NL) question. 
(3)~The rephrased questions are further validated to ensure that they share the same meaning as the original template questions.

\begin{figure*}[!tp]
\vspace{-4mm}
\subfigure[]{\includegraphics[width=0.63\textwidth]{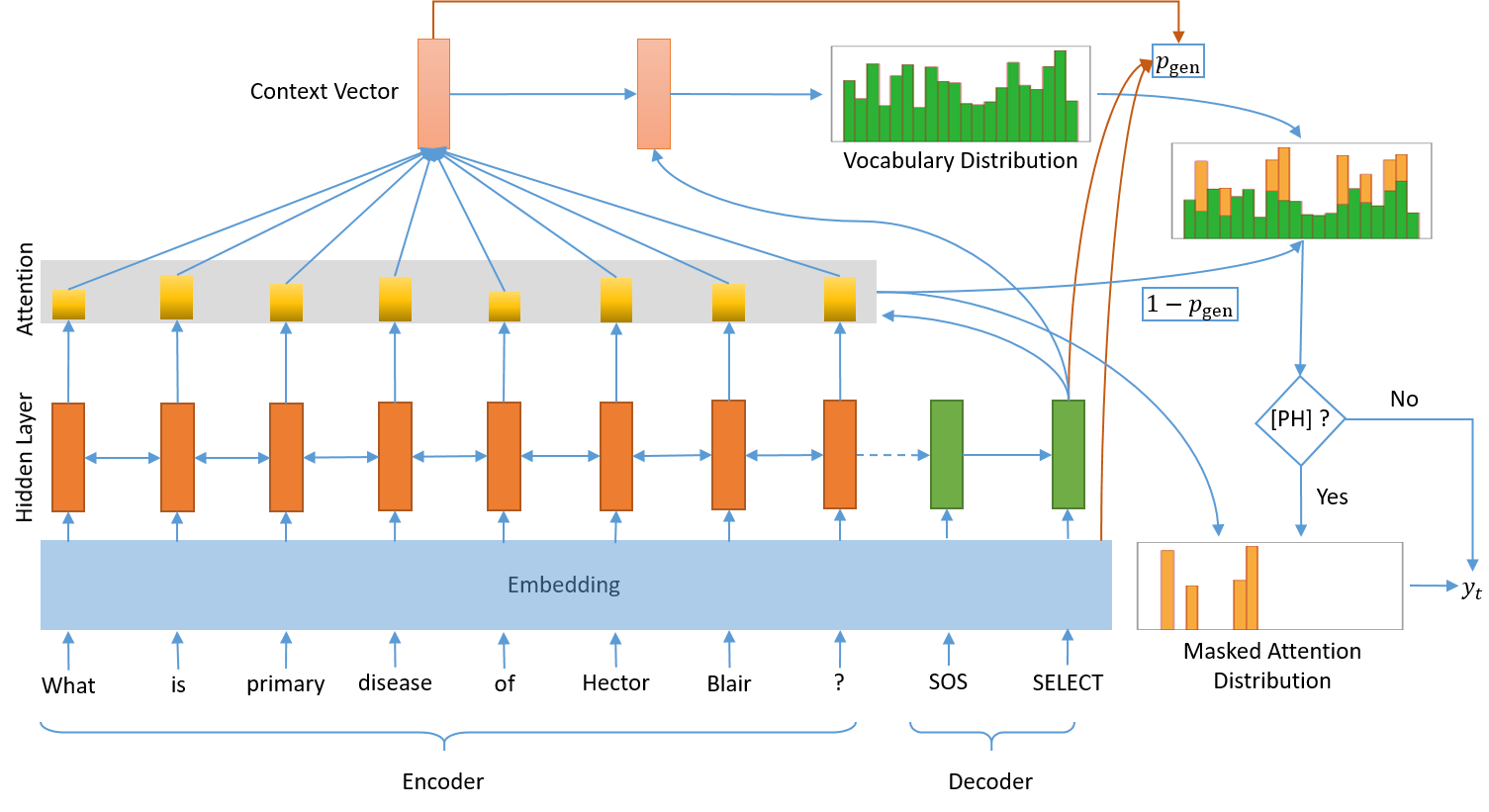}}
\hspace{2mm}
\subfigure[]{\includegraphics[width=0.35\textwidth]{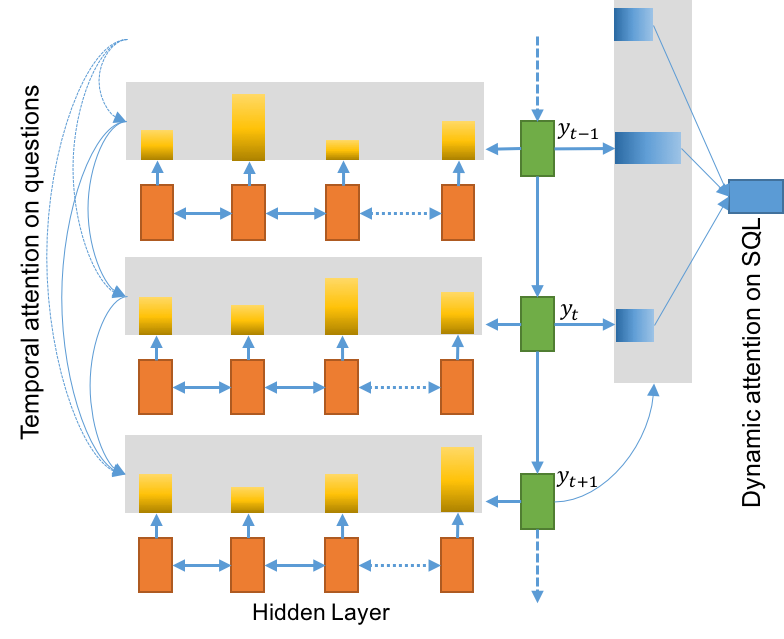}}
\vspace{-6mm}
\caption{(a) The overall framework of the proposed TREQS model. [PH] represents the out of vocabulary words in condition values. (b) Illustration of dynamic and temporal attention mechanisms used in TREQS.}
\vspace{-3mm}
\label{fig:model}
\end{figure*}

\subsection{MIMICSQL Statistics}
MIMICSQL dataset is publicly available at\footnote{https://github.com/wangpinggl/TREQS}.
We include $10,000$ Question-SQL pairs in MIMICSQL whose basic statistics are provided in Figure~\ref{fig:stat2} and Table~\ref{tab:stat}. 
Figure~\ref{fig:stat2}(a) and Figure~\ref{fig:stat2}(b) shows the distributions of the question length for template questions and natural language questions, respectively. The distribution of the SQL length is given in Figure~\ref{fig:stat2}(c).
Figure~\ref{fig:stat2}(d) shows the distribution of number of questions over five tables. Note that the total number of questions in Figure~\ref{fig:stat2}(d) is more than $10,000$ since some questions are related to more than one table.
\vspace{-2mm}
\section{A Translate-Edit Model for Question-to-SQL query generation}
\label{sec4}
In this section, we will first formulate the Question-to-SQL query generation problem.
Then, we present our TREQS model in detail.

\vspace{-2mm}
\subsection{Problem Formulation}
In this paper, we aim to translate healthcare related questions asked by doctors to database queries and then retrieve the answer from health records.
We adapt the language generation approach in our model, since questions may be related to a single table or multiple tables, and keywords in the questions may not be accurate due to the healthcare terminology involved.
To tackle the challenges for general applications, we propose a translate-edit model that first generates a query draft using a language generation model and then edits based on the table schema.

Let us denote a given question by $x=(x_1, x_2, ..., x_J)$, the table schema context information as $z$ and the corresponding query as $y=(y_1, y_2, ..., y_T)$, where $J$ and $T$ represents the length of the input and output, respectively. $x_j$ and $y_t$ denote the one-hot representations of the tokens in the question and query, respectively.
Then, the goal of our model is to infer $y$ from $x$ based on $z$ with probability $P(y|x,z)$.
In our approach, we assume that the table schema information $z$ is implicitly included in the input questions as semantic information.
Therefore, during the translation, we only need to deal with inferring $y$ from $x$.
However, since the exact table schema has not appeared at this stage, the generated query can only roughly capture this information.
At the second stage, we edit the query draft based on the table schema and look-up tables of content keywords to recover the exact information.
This two-stage strategy allows us to easily adapt our model to other general purpose tasks.
In the following sections, we will introduce our model layer-by-layer in more detail.

\vspace{-4mm}
\subsection{The Proposed TREQS Model}
Now we introduce the details of the three components in the proposed \textbf{TR}anslate-\textbf{E}dit Model for \textbf{Q}uestion-to-\textbf{S}QL~(\textbf{TREQS}) generation.
Figure~\ref{fig:model}(a) shows the framework of the proposed model.

\subsubsection{Sequence-to-Sequence Framework}
\label{sec:seq2seq}

We adopt a RNN sequence-to-sequence (Seq2Seq) framework for the Question-to-SQL generation task.
Our Seq2Seq framework is composed of a question encoder (a single-layer bidirectional LSTM \cite{hochreiter1997long}) and a SQL decoder (a single-layer unidirectional LSTM).
The encoder reads a sequence of word embeddings of input tokens and turns them into a sequence of encoder hidden states (features) $h^e=(h^e_1, h^e_2, \dots, h^e_J)$, where the superscript $e$ indicates that the hidden states are obtained from the encoder, and $h^e_j=\vec{h_j^e}\oplus\cev{h}_{J-j+1}^e$ is the concatenation of the hidden states of forward and backward LSTM.
At each decoding step $t$, the decoder takes the encoder hidden states and word embedding of the previous token as an input and produce a decoder hidden state $h^d_t$. 
Both word embeddings in the encoder and decoder are taken from the same matrix $W_\text{emb}$.
The decoder LSTM hidden and cell states are initialized with\vspace{-0.06in}
\begin{equation}
\aligned
&h^d_0=\tanh\left(W_\text{e2dh}\big(\vec{h}^e_{J}\oplus\cev{h}^e_{1}\big)+b_\text{e2dh}\right)\\
&c^d_0=\tanh\left(W_\text{e2dc}\big(\vec{c}^e_{J}\oplus\cev{c}^e_{1}\big)+b_\text{e2dc}\right)
\endaligned
\end{equation}
where the weight matrices $W_\text{e2dh}$, $W_\text{e2dc}$, and vectors $b_\text{e2dh}$, $b_\text{e2dc}$ are learnable parameters.

\subsubsection{Temporal Attention on Question}
At each decoding step $t$, the decoder not only takes its internal hidden state and previously generated token as input, but also selectively focuses on parts of the question that are relevant to the current generation.
However, the standard attention models proposed in the literature \cite{bahdanau2014neural,luong2015effective} cannot prevent the decoder from repetitively attending on the same part of the question, therefore, we adopt a temporal attention strategy \cite{nallapati2016abstractive} that was demonstrated to be effective in tackling such problem. 

To achieve this goal, we first define an alignment score function between the current decoder hidden state and each of the encoder hidden states as follows:\vspace{-0.06in}
\begin{equation}
s^e_{tj}=(h_j^e)^\top W_\text{align}h_t^d
\end{equation}
where $W_\text{align}$ are parameters.
As shown in the left-hand side of Figure~\ref{fig:model}(b), to avoid repetitive attention, we penalize the tokens that have obtained high attention scores in the previous decoding steps with the following normalization rule:\vspace{-0.06in}
\begin{eqnarray}
s^\text{temp}_{tj}=
\begin{cases}
\exp{(s^{e}_{tj})} & \text{if } t=1 \\
\frac{\exp{(s^e_{tj})}}{\sum_{k=1}^{t-1}\exp{(s^e_{kj})}} & \text{if } t > 1
\end{cases},\ \ \ 
\alpha_{tj}=\frac{s^\text{temp}_{tj}}{\sum_{k=1}^{J}s^\text{temp}_{tk}}
\label{eqn:temporal}
\end{eqnarray}
where $s_{tj}^\text{temp}$ is the new alignment score with temporal dependency, and $\alpha_{tj}$ is an attention weight at current decoding step.
With the temporal attention mechanism, we finally obtain a context vector for the input question as follows:\vspace{-0.06in}
\begin{equation}
z_t^e=\sum_{j=1}^{J}\alpha_{tj}h_j^e.
\end{equation}

\vspace{-2mm}
\subsubsection{Dynamic Attention on SQL}
In our Question-to-SQL generation task, different parts of a query may not strictly have sequential dependency.
For example, switching two conditions in a query will yield the same query.
However, when generating the condition values, the decoder may need to not only take the previously generated token, its own hidden states and encoder context vector into consideration, but also places more attention on the previously generated table names and headers as shown in the right-hand side of Figure~\ref{fig:model}(b).
Therefore, we introduce a dynamic attention mechanism to the decoder \cite{shi2019leafnats, shi2018neural}, which allows it to dynamically attend on the previous generated tokens.

More formally, for $t>1$, the alignment scores (denoted by $s^d_{t\tau}$, $\tau\in\{1,..., t-1\}$) on previously generated tokens can be calculated in the same manner as the alignment scores for the encoder.
Then, the attention weight for each token is calculated as follows:\vspace{-0.06in}
\begin{equation}
\alpha^d_{t\tau}=\frac{\exp(\text{s}^d_{t\tau})}{\sum_{k=1}^{t-1}\exp(\text{s}^d_{tk})}
\end{equation}
With the attention distribution and the decoder hidden states, we can calculate the decoder-side context vector as follows:\vspace{-0.06in}
\begin{equation}
z_t^d=\sum_{\tau=1}^{t-1}\alpha^d_{t\tau}h^d_{\tau}
\end{equation}

\vspace{-2mm}
\subsubsection{Controlled Generation and Copying}
A Question-to-SQL generation task is very different from the general purpose language generation tasks.
First, there are strict templates for SQL queries.
For example, SELECT \$AGG\_OP (\$AGG\_COLUMN)$^+$ FROM \$TABLE WHERE (\$COND\_COLUMN \$COND\_OP \$COND\_VAL)$^+$ is the template we used.
Second, the aggregation and condition columns in queries are table headers, which usually do not exactly appear in the questions.
For instance, for a given question: ``How many patients who have bowel obstruction and stay in hospital for more than 10 days?'', its corresponding query looks like ``SELECT COUNT ( PATIENT\_ID ) FROM DEMOGRAPHIC WHERE PRIMARY\_DISEASE = bowel obstruction AND DAYS\_OF\_STAY $>$ 10''.
Obviously, we cannot find words, like PATIENT\_ID, PRIMARY\_DISEASE, and DAYS\_OF\_STAY, in the question.
Third, the values of conditions should be best possibly retrieved from questions, such as ``bowel obstruction" and ``10" in the above example,
since the questions may contain terms that are out-of-vocabulary (OOV).

Because of these characteristics, our decoder combines a generation network and a pointer network \cite{vinyals2015pointer} for the token generation.
The pointer network has been widely used in language modeling and generation tasks, such as abstractive text summarization \cite{see2017get} and question-answering \cite{mccann2018natural}, due to its ability of copying OOV tokens in the source and context sequences to the target sequences.
However, in our model, it is primarily used for generating the words in-vocabulary and putting placeholders, denoted as [PH], for OOV words. 
Intrinsically, it is only used in generating condition values in SQL queries.
Formally, to generate a token at step $t$, we first calculate the probability distribution on a vocabulary $\mathcal{V}$ as follows:
\vspace{-0.04in}
\begin{equation}
\aligned
    \tilde{h}_t^d&=W_z\big(z_t^e\oplus z_t^d\oplus h_t^d\big)+b_z \\
    P_{\mathcal{V}, t} &= \text{softmax}\left(W_\text{emb}(W_\text{d2v}\tilde{h}_t^d+b_\text{d2v})\right)
    \label{eqn:prob_vocab}
\endaligned
\end{equation}
where $W_z$, $W_\text{d2v}$, $b_z$, and $b_\text{d2v}$ are parameters.
We reuse the syntactic and semantic information contained in the word embedding matrix in token generation.
Then, combining with the pointer mechanism, the probability of generating a token $y_t$ is calculated by\vspace{-0.06in}
\begin{equation}
P(y_t)=p_{\text{gen},t}P_\text{gen}(y_t)+(1-p_{\text{gen},t})P_\text{ptr}(y_t)
\end{equation}
where the probability $P_\text{gen}(y_t)$ given by the generation network is calculated as follows:\vspace{-0.06in}
\begin{equation}
P_\text{gen}(y_t)=\begin{cases}
P_{\mathcal{V},t}(y_t) & y_t\in\mathcal{V}\\
0& \text{otherwise}
\end{cases}
\end{equation}
The probability $P_\text{ptr}(y_t)$ by the pointer network is obtained with the following attention distribution\vspace{-0.06in}
\begin{equation}
P_\text{ptr}(y_t)=\begin{cases}
\sum_{j:x_j=y_t}\alpha^e_{tj} & y_t\in\mathcal{X}\cap\mathcal{V}\\
0 & \text{otherwise}
\end{cases}
\end{equation}
where $\mathcal{X}$ is a set with all tokens in a question. $p_{\text{gen},t}$ is a `soft-switch' (probability) of using a generation network for token generation\vspace{-0.06in}
\begin{equation}
p_{\text{gen},t}=\sigma(W_\text{gen}z_t^e\oplus h_t^d\oplus E_{y_{t-1}}+b_\text{gen})
\label{eqn:pgen}
\end{equation}
where $E_{y_{t-1}}$ is the word embedding of the previous token $y_{t-1}$. $W_\text{gen}$ and $b_\text{gen}$ are model parameters.
Note that all OOV words in the question have been replaced with the placeholder [PH] for the condition values.
In our model, the vocabulary is a union of two sets, i.e., vocabulary of regular tokens and a vocabulary of template keywords as well as table names and headers, denoted as $\mathcal{V}_\text{schema}$.
Since $\mathcal{X}\cap\mathcal{V}_\text{schema}=\emptyset$, the template, table names and headers in a SQL rely only on the generation network.
On the other hand, keywords of the condition values and placeholder are obtained from both generation and pointer networks.
Note that we always switch the option of [PH] in Figure~\ref{fig:model}(a) to``No" during training.

With the final probability of generating a token $y_t$, we are ready to define our loss function.
In this paper, we adopt the cross-entropy loss which tries to maximize the log-likelihood of observed sequences (ground-truth), i.e.,
\vspace{-2mm}
\begin{equation}
    \mathcal{L}=-\log P_\theta(\hat{y}|x)=\sum_{t=1}^{T}\log P_\theta(\hat{y}_t|\hat{y}_{<t},x)
\end{equation}
\vspace{-0.5mm}
where $\theta$ denotes all the model parameters, including weight matrices $W$ and biases $b$. $\hat{y}=(\hat{y}_1,\hat{y}_2,...,\hat{y}_{T})$ represents a ground-truth SQL sequence in the training data and
$\hat{y}_{<t}=(\hat{y}_1,\hat{y}_2,...,\hat{y}_{t-1})$.
\vspace{-2mm}

\subsubsection{Placeholder Replacement}
After a query has been generated, we replace each [PH] with a token in the source question.
For a [PH] at time step $t'$, the replacement probability is calculated by\vspace{-0.06in}
\begin{equation}
P_\text{rps}(y_{t'})=\begin{cases}
\sum_{j:x_j=y_{t'}}\alpha^e_{t'j} & y_{t'}\in\mathcal{X}-\mathcal{V}\\
0 & \text{otherwise}
\end{cases}
\end{equation}
Here, we implement this technique by applying a mask (0 or 1) on the attention weights (named as masked attention mechanism). 
This replacement technique can make use of the semantic relationships (captured by attention and decoder LSTM) between previously generated words and their neighboring OOV words.
Intuitively, if the model attends word $x_j$ at the step $t-1$, it has a high chance of attending the neighboring words of $x_j$ at step $t$.
This meta-algorithm can be used for any attention-based Seq2Seq model.

\vspace{-2mm}
\subsubsection{Recover Condition Values with Table Content}
So far, we have used our translate-edit model to translate given questions on a table to the SQL queries without explicitly using any table content and schema.
However, we cannot guarantee that all these queries are executable, since the condition values in the questions may not be accurate.
In the aforementioned example, the doctor may ask
``How many patients who have \textbf{bowel obstruct} and stay in hospital for more than 10 days?", then, one of the conditions in the SQL is ``PRIMARY\_DISEASE = \textbf{bowel obstruct}".
Obviously, we will get a different answer since 
\textbf{bowel obstruct} does not appear in the database.
To alleviate this problem, \textit{we propose a condition value recover technique to retrieve the exact condition values} based on the predicted ones.
This approach makes use of string matching metric ROUGE-L \cite{lin2004rouge} (L denotes the longest common sub-sequence) to find the most similar condition value from the look-up table for each predicted one, and then replaces it.
In our implementation, we calculate both word- and character-level similarities, i.e., ROUGE-L scores, between two sequences.
\vspace{-2mm}
\section{Experiments}
\label{sec5}
In this section, we first introduce the datasets used in our experiments, and then briefly describe the baseline comparison methods, implementation details, and evaluation metrics. 
Finally, different sets of qualitative and quantitative results are provided to analyze the query generation performance of the proposed model.  
\vspace{-1mm}

\subsection{Experimental Settings}
\subsubsection{Dataset Description}
\label{sec5-1}
We use both template and natural language (NL) questions in MIMICSQL dataset (described in Section~\ref{sec3}) for evaluation. We first tokenize both source questions and target SQL queries using Spacy package\footnote{https://spacy.io/}. 
Then, they are randomly split into training, development and testing sets in the ratio of $0.8/0.1/0.1$.
To recover the condition values, we also created a look-up table that contains table schema and keywords, i.e., table name, header and keywords of each column. Finally, for template questions in the testing set, we also generated a dataset that has missing information and typos (\textit{testing with noise}) to demonstrate the effectiveness of our condition value recover technique.

\vspace{-2mm}
\subsubsection{Comparison Methods}
We demonstrate the superior performance of our TREQS model by comparing it with the following methods. The first two are slot filling methods and generate logic format of queries, while the others produce SQL queries directly.

\begin{itemize}[leftmargin=*]
\item \textbf{Coarse2Fine model~\cite{dong2018coarse}:} It is a two-stage structure-aware neural architecture for semantic parsing. For a given question, a rough sketch of the logical form is first generated by omitting low-level information, such as the arguments and name entities, which will be filled in the second step by considering both the natural language input and the generated sketch. 

\item \textbf{Multi-table SQLNET (M-SQLNET)~\cite{xu2017sqlnet}:} For SQL with multiple conditions, it may have multiple equivalent variants by varying the order of conditions. 
SQLNET mainly focuses on tackling the unordered property by leveraging the structure-based dependencies in SQL. 
However, it can only handle questions on a single table under the table-aware assumption. In this paper, we implemented a multi-table version of SQLNET for comparison.

\item \textbf{Sequence-to-Sequence (Seq2Seq) model \cite{luong2015effective}:}
In this model, there is a bidirectional LSTM encoder and a LSTM decoder. To be consistent with this paper, we adopt the ``general'' global attention mechanism described in \cite{luong2015effective}. The placeholder replacement algorithm is also used in the query generation step to tackle the OOV words problem in this model.

\item \textbf{Pointer-Generator Network (PtrGen) \cite{see2017get}:} 
The pointing mechanism is primarily used to deal with the OOV words.
Therefore, an extended vocabulary of all OOV words in a batch is built at each training step to encourage the copying of low-frequency words in the source questions, which is different from our model.
In our pointer network, we encourage the model to either copy tokens related to the condition values or put placeholders.
\end{itemize}

Note that the proposed condition value recover mechanism can be combined with different models that directly generate SQL queries, therefore, we also apply it to the results obtained from Seq2Seq and PtrGen to boost their performance. 
However, it is not applicable to Coarse2Fine and M-SQLNET since their predicted condition values have already been in the look-up table.
\vspace{-2mm}

\subsubsection{Implementation Details}
\label{implement}
We implemented the proposed TREQS model and M-SQLNET with Pytorch~\cite{paszke2017automatic}.
For all language generation models, the dimension of word embeddings and the size of hidden states (both encoder and decoder hidden states) are set to be 128 and 256, respectively. Instead of using pre-trained word embeddings \cite{pennington2014glove}, we learn them from scratch. ADAM optimizer~\cite{kingma2014adam} with hyper-parameter $\beta_1 =0.9$, $\beta_2 =0.999$ and $\epsilon=10^{-8}$ is adopted to train the model parameters.
The learning rate is set to be $0.0005$ with a decay for every $2$ epochs and gradient clipping is used with a maximum gradient norm of 2.0.
During the training, we set the mini-batch size to be 16 in all our experiments and run all models for $20$ epochs.
The development set is used to determine the best model parameters.
During the testing, we implement a beam search algorithm for the SQL generation and the beam size is set to be 5.
To build the vocabulary, we keep the words with a minimum frequency of 5 in the training set.
Thus, the vocabulary size is $2353$ and it is shared between the source question and target SQL.
In our experiments, both the source questions and SQL queries are truncated to $30$ tokens. The implementation of our proposed TREQS method is made publicly available at\footnote{https://github.com/wangpinggl/TREQS}.

\vspace{-2mm}
\subsubsection{Evaluation Metrics}
To evaluate the performance of different Question-to-SQL generation models, we mainly adopt the following two commonly used evaluation metrics~\cite{zhong2017seq2sql}. (1) \textbf{Execution accuracy} is defined as $Acc_{EX}={N_{EX}}/{N}$, where $N$ denotes the number of Question-SQL pairs in MIMICSQL, and $N_{EX}$ represents the number of generated SQL queries that can result in the correct answers~\cite{zhong2017seq2sql}. Note that execution accuracy may include questions that are generated with incorrect SQL queries which lead to correct query results. (2) In order to overcome the disadvantage of execution accuracy, \textbf{logic form accuracy}~\cite{zhong2017seq2sql}, defined as $Acc_{LF}={N_{LF}}/{N}$, is commonly used to analyze the string match between the generated SQL query and the ground truth query. Here, $N_{LF}$ denotes the number of queries that match exactly with the ground truth query.

\begin{table}[!tp]
\centering
\caption{The SQL prediction performance results using logic form accuracy ($Acc_{LF}$) and execution accuracy ($Acc_{EX}$).}
\vspace{-3mm}
\renewcommand\arraystretch{1.1}
\setlength\tabcolsep{1.0pt}
\resizebox{1.0\linewidth}{!}{
\begin{tabular}{|l|c|c|c|c|c|c|c|c|}\hline
& \multicolumn{4}{c|}{\textbf{Template Questions}} & \multicolumn{4}{c|}{\textbf{NL Questions}}  \\\hline
{\textbf{Method}} & \multicolumn{2}{c|}{\textbf{Development}} &\multicolumn{2}{c|}{\textbf{Testing}} & \multicolumn{2}{c|}{\textbf{Development}} & \multicolumn{2}{c|}{\textbf{Testing}} \\\hline
&$Acc_{LF}$ & $Acc_{EX}$ & $Acc_{LF}$ & $Acc_{EX}$ &$Acc_{LF}$ & $Acc_{EX}$ &$Acc_{LF}$ & $Acc_{EX}$\\\hline
Coarse2Fine & 0.298 & 0.321 & 0.518 & 0.526 &0.217 & 0.309 & 0.378 & 0.496 \\\hline
M-SQLNET & 0.258 & 0.588 & 0.382 & 0.603 &0.086 &0.225  &0.142 &0.260   \\\hline
Seq2Seq & 0.098 & 0.372 & 0.160 & 0.323 &0.076 &0.112  &0.091 &0.131   \\\hline
Seq2Seq + recover &0.138 &0.429 &0.231 &0.397 &0.092 &0.195  &0.103 &0.173  \\\hline

PtrGen & 0.312 & 0.536 & 0.372 & 0.506 &0.126 &0.174  & 0.160 &0.222  \\\hline
PtrGen + recover &0.442 &0.645 &0.426 &0.554 &0.181 &0.325  &0.180 &0.292  \\\hline

TREQS (ours) &\underline{0.712} &\underline{0.803} &\underline{0.802} &\underline{0.825} &\underline{0.451} &\underline{0.511} &\underline{0.486} &\underline{0.556}  \\\hline
TREQS + recover &\textbf{0.853} &\textbf{0.924} &\textbf{0.912} &\textbf{0.940} &\textbf{0.562} &\textbf{0.675} &\textbf{0.556} &\textbf{0.654} \\\hline 
\end{tabular}
} 
\label{tab:results1} 
\vspace{-5.5mm}
\end{table}

\vspace{-2mm}
\subsection{Experimental Results}
\subsubsection{Query Generation Performance}

Table~\ref{tab:results1} provides the quantitative results on both template questions and NL questions for different methods. The best performing methods are highlighted in bold and the second best performing methods are underlined. It can be observed from Table~\ref{tab:results1} that the Seq2Seq model is the worst performer among all the compared methods due to its poor generating behavior, including factual errors, repetitions and OOV words. PtrGen performs significantly better than Seq2Seq model since it is able to copy words from the input sequence to the target SQL.
As seen from the results, it can capture the factual information and handle OOV words more efficiently.
It works well when most words in the target sequence are copied from the source sequence, which is similar to other problems such as abstractive text summarization task \cite{see2017get,gehrmann2018bottom}.
However, in Question-to-SQL task, most tokens (template, table names and headers) are obtained from generation and only condition values are copied from questions to queries. 
Therefore, the task discourages copying in general, which causes PtrGen model to produce the condition values by generation instead of copying, thus increasing the chances of making mistakes.
Coarse2Fine achieves outstanding performance for the questions on a single table.
The limitation of Coarse2Fine is that it cannot handle complex SQL generation, such as queries including multiple tables.
However, it still outperforms both Seq2Seq and PtrGen in most of the cases.
Compared to Coarse2Fine, the M-SQLNET method considers the dependencies between slots using a dependency graph determined by the intrinsic structure of SQL. 
It performs significantly better than Seq2Seq and PtrGen on both testing and testing with noise set (in Table \ref{tab:results-noise}). 
It also significantly outperforms Coarse2Fine based on the execution accuracy.
Compared to all the aforementioned baseline methods, our proposed TREQS model gains a significant performance improvement on both development and testing dataset and $30$ percent, on average, more accurate than others.

We have also applied the proposed condition value recover technique to three language generation models.
It can be observed that such a heuristic approach can significantly boost the performance of these models.
From our experiments, we found that language models fail in many cases because they cannot capture all keywords of condition values.
As a result, they are not executable or may yield different answers.
Hence, the recover mechanism can correct these errors in the conditions of SQL by making the best use of the look-up table.
Moreover, as shown in Table~\ref{tab:results-noise}, after applying some noise to the template testing questions by removing partial condition values or using abbreviations of words, the performance of different models drops.
Our TREQS model is affected significantly because it strongly relies on the pointing mechanism to copy keywords of condition values from questions to queries.
However, as we can see, the recover mechanism can still correct most of the errors, thus improving the accuracy by more than $20\%$, which is $13\%$ for the testing set without introducing noise.

\begin{table}[!tp]
\centering
\caption{The SQL prediction performance results and their break-down on \textbf{template testing questions with noise}.}
\vspace{-3mm}
\renewcommand\arraystretch{1.25}
\setlength\tabcolsep{1.0pt}
\resizebox{1.0\linewidth}{!}{
\begin{tabular}{|l|c|c|c|c|c|c|c|c|}\hline
{\textbf{Method}} & \multicolumn{2}{c|}{\textbf{Overall}} &\multicolumn{6}{c|}{\textbf{Break-down}} \\\hline
&$Acc_{LF}$ & $Acc_{EX}$ &$Agg_{op}$ & $Agg_{col}$ & $Table$ & $Con_{col+op}$ &$Con_{val}$ & Average\\\hline
Coarse2Fine & 0.444 & 0.526 &0.528 &0.528 &0.528 &0.520 &0.444 &0.510  \\\hline
M-SQLNET  & 0.356 & 0.606  & \textbf{1.000} &0.953 &\textbf{0.998}	&0.875	&0.376	&0.840 \\\hline
Seq2Seq  & 0.157 & 0.320 & 0.997 &0.862	&0.967	&0.817	&0.206	&0.770   \\\hline
Seq2Seq + recover  &0.225 &0.389 & \underline{0.999} &0.862	&0.967	&0.817	&0.290	&0.787 \\\hline

PtrGen & 0.301 & 0.451 & \underline{0.999}	&\underline{0.988}	&0.991	&\underline{0.970}	&0.309	&0.851  \\\hline
PtrGen + recover  &0.353 &0.498 &\underline{0.999}	&\underline{0.988}	&0.991 &\underline{0.970} &0.360 &0.862  \\\hline

TREQS (ours)  &\underline{0.699} &\underline{0.756}  &\textbf{1.000} &\textbf{0.996} &\underline{0.995}	&\textbf{0.976}	&\underline{0.706}	&\underline{0.935}  \\\hline
TREQS + recover  &\textbf{0.872} &\textbf{0.907} &\textbf{1.000} &\textbf{0.996} &\underline{0.995} &\textbf{0.976}	&\textbf{0.877} &\textbf{0.969} \\\hline 
\end{tabular}
} 
\label{tab:results-noise} 
\vspace{-4.5mm}
\end{table}

\vspace{-1.5mm}
\subsubsection{Break-down Generation Performance}
In order to further evaluate the performance on each component of SQL query, in Tables~\ref{tab:results-noise}, ~\ref{tab:results-2} and~\ref{tab:results-3}, we provide the break-down accuracy results based on SQL query structure, including aggregation operation ($Agg_{op}$), aggregation column ($Agg_{col}$), table ($Table$), condition column along with its operation ($Con_{col+op}$), and condition value ($Con_{val}$). 
The results of Coarse2Fine are not provided due to its table-aware assumption and its inability in handling multi-table questions. We can observe that there is no significant difference between these methods on predictions of both aggregation operation and table. Seq2Seq model performs relatively worse on aggregation column and condition column and its operation.

\begin{table*}[!tp]
\centering
\caption{Accuracy of break-down matching on \textbf{template questions} in MIMICSQL dataset.}
\vspace{-3mm}
\renewcommand\arraystretch{1.0}
\setlength\tabcolsep{4pt}
\begin{tabular}{|l|c|c|c|c|c|c|c|c|c|c|c|c|}\hline
{\textbf{Method}} & \multicolumn{6}{c|}{\textbf{Development}} &\multicolumn{6}{c|}{\textbf{Testing}}\\\cline{1-13}
&$Agg_{op}$ & $Agg_{col}$ & $Table$ & $Con_{col+op}$ &$Con_{val}$ & Average &$Agg_{op}$ & $Agg_{col}$ & $Table$ & $Con_{col+op}$ &$Con_{val}$ & Average \\\hline
Coarse2Fine &0.321  &0.321 &0.321 &0.321 &0.298 &0.316 &0.528 &0.528 &0.528 &0.520 &0.518 &0.524  \\\hline
M-SQLNet &\textbf{1.000} &0.978 &\underline{0.994} &0.876 &0.274 &0.824 &\textbf{1.000}  &0.956 &\textbf{0.996} &0.881 &0.401 &0.847  \\\hline
Seq2Seq &\underline{0.999} &0.950 &0.972 &0.761 &0.119  &0.760 &0.999 &0.865 &0.963 & 0.818 &0.210 & 0.771  \\\hline
Seq2Seq + recover &\underline{0.999} & 0.950 &0.972 &0.761 &0.163 &0.769 &0.999 &0.865 &0.963 &0.818 &0.296 &0.788  \\\hline
PtrGen &\underline{0.999} &\underline{0.991} &0.992 &\textbf{0.979} &0.325 &0.857 &\textbf{1.000} &0.988 &\underline{0.992} &\textbf{0.985} &0.381 & 0.869 \\\hline
PtrGen + recover &\underline{0.999} &\underline{0.991} &0.992 &\textbf{0.979} &0.449 &0.882 &\textbf{1.000} &0.988 &\underline{0.992} &\textbf{0.985} &0.433 &0.880  \\\hline

TREQS (ours) &\textbf{1.000} &\textbf{0.999} &\textbf{0.995} &\underline{0.924} &\underline{0.719} &\underline{0.927} &\textbf{1.000}  &\underline{0.995} &\textbf{0.996} &0.980 &\underline{0.810} &\underline{0.956} \\\hline
TREQS + recover &\textbf{1.000} &\textbf{0.999} &\textbf{0.995} &\underline{0.924} &\textbf{0.859} &\textbf{0.955} &\textbf{1.000} &\textbf{0.996} &\textbf{0.996} &\underline{0.984} &\textbf{0.918} &\textbf{0.979} \\\hline
\end{tabular}
\vspace{-1.5mm}
\label{tab:results-2} 
\end{table*}

\begin{table*}[!tp]
\centering
\caption{Accuracy of break-down matching on \textbf{NL questions} in MIMICSQL dataset.}
\vspace{-3mm}
\renewcommand\arraystretch{1.0}
\setlength\tabcolsep{4pt}
\begin{tabular}{|l|c|c|c|c|c|c|c|c|c|c|c|c|}\hline
{\textbf{Method}} & \multicolumn{6}{c|}{\textbf{Development}} &\multicolumn{6}{c|}{\textbf{Testing}}\\\cline{1-13}
&$Agg_{op}$ & $Agg_{col}$ & $Table$ & $Con_{col+op}$ &$Con_{val}$ & Average &$Agg_{op}$ & $Agg_{col}$ & $Table$ & $Con_{col+op}$ &$Con_{val}$ & Average \\\hline
Coarse2Fine &0.319  &0.313 &0.321  &0.260 &0.214 &0.285  &0.524 & 0.490 &0.528 & 0.448 &0.413 & 0.481\\\hline
M-SQLNet &\textbf{0.994}  &\textbf{0.939} &0.933  &0.722 &0.080 &0.734  &\underline{0.989} &\textbf{0.873}  &\textbf{0.941} & 0.749 &0.140 &0.738 \\\hline
Seq2Seq &0.978  &0.872 &0.926  &0.466 &0.137 &0.676  &0.970  &0.696  &0.892 &0.563  &0.239 &0.672  \\\hline
Seq2Seq + recover &0.978  &0.872 &0.926  &0.471 &0.174 &0.684  &0.970 &0.696  &0.892 &0.565  &0.296 &0.684\\\hline
PtrGen &0.987  &\underline{0.917} &\textbf{0.944}  & \underline{0.795} &0.172 &0.766  &0.987 &\underline{0.830}  &0.926 &0.824  &0.214 &0.757 \\\hline
PtrGen + recover &0.987  &\underline{0.917} &\textbf{0.944} &\underline{0.795} & 0.236 &0.776  &0.987 &\underline{0.830}  &0.926 & 0.824 & 0.235 &0.760\\\hline
TREQS (ours) &\underline{0.990} &0.912 &\underline{0.942} &\textbf{0.834} &\textbf{0.574} &\textbf{0.850} &\textbf{0.993} &0.827 &\textbf{0.941} &\underline{0.841} &\underline{0.679} &\underline{0.856} \\\hline
TREQS + recover &\underline{0.990} & 0.912 &\underline{0.942} &\textbf{0.834} &\textbf{0.694} &\textbf{0.873} &\textbf{0.993} &0.827 &\textbf{0.941} &\textbf{0.844} &\textbf{0.763} & \textbf{0.874}\\\hline
\end{tabular}
\label{tab:results-3} 
\end{table*}

\begin{table*}[!hp]
\centering
\caption{SQL Queries generated by different models on two \textbf{ NL questions} in testing set. The incorrectly predicted words are highlighted in red color.}
\vspace{-3mm}
\renewcommand\arraystretch{0.9}
\resizebox{\linewidth}{!}{
\begin{tabular}{|m{7em}|m{27em}|m{27em}|}\hline
\multicolumn{1}{|c|}{\textbf{Method}} &\multicolumn{1}{c|}{\textbf{Example 1}} &\multicolumn{1}{c|}{\textbf{Example 2}}\\\hline\hline
\textbf{Question} &how many female patients underwent the procedure of abdomen artery incision?  &how many patients admitted in emergency were tested for ferritin?  \\\hline

\textbf{Ground truth} &\textbf{select} count (distinct demographic."subject\_id") \textbf{from} demographic inner join procedures on demographic.hadm\_id = procedures.hadm\_id \textbf{where} demographic."gender" = "f" and procedures."short\_title" = "abdomen artery incision"  & \textbf{select} count (distinct demographic."subject\_id") \textbf{from} demographic inner join lab on demographic.hadm\_id = lab.hadm\_id \textbf{where} demographic."admission\_type" = "emergency" and lab."label" = "ferritin" \\\hline

M-SQLNET &\textbf{select} count (distinct demographic."subject\_id") \textbf{from} demographic inner join procedures on demographic.hadm\_id = procedures.hadm\_id \textbf{where} demographic."gender" = "f" and procedures."short\_title" = "\textcolor{red}{parent infus nutrit sub}"  &\textbf{select} count (distinct demographic."subject\_id") \textbf{from} demographic inner join lab on demographic.hadm\_id = lab.hadm\_id \textbf{where} demographic."admission\_type" = "emergency" and lab."label" = "\textcolor{red}{po2}" \\\hline

Seq2Seq &\textbf{select} count (distinct demographic."subject\_id") \textbf{from} demographic inner join procedures on demographic.hadm\_id = procedures.hadm\_id \textbf{where} demographic."gender" = "\textcolor{red}{m}" and \textcolor{red}{procedures."long\_title"} = "\textcolor{red}{other abdomen}"  & \textbf{select} count (distinct demographic."subject\_id") \textbf{from} demographic inner join lab on demographic.hadm\_id = lab.hadm\_id \textbf{where} \textcolor{red}{demographic."admission\_location"} = "\textcolor{red}{phys referral/normal deli}" and \textcolor{red}{lab."itemid"} = "ferritin"\\\hline

Seq2Seq+recover &\textbf{select} count (distinct demographic."subject\_id") \textbf{from} demographic inner join procedures on demographic.hadm\_id = procedures.hadm\_id \textbf{where} demographic."gender" = "\textcolor{red}{m}" and \textcolor{red}{procedures."long\_title"} = "\textcolor{red}{other bronchoscopy}"  &\textbf{select} count (distinct demographic."subject\_id") \textbf{from} demographic inner join lab on demographic.hadm\_id = lab.hadm\_id \textbf{where} \textcolor{red}{demographic."admission\_location"} = "\textcolor{red}{phys referral/normal deli}" and \textcolor{red}{lab."itemid"} = "\textcolor{red}{51200}"  \\\hline

PtrGen &\textbf{select} count (distinct demographic."subject\_id") \textbf{from} demographic inner join procedures on demographic.hadm\_id = procedures.hadm\_id \textbf{where} demographic."gender" = "f" and \textcolor{red}{procedures."long\_title"} = "\textcolor{red}{spinal abdomen artery}"  &\textbf{select} count (distinct demographic."subject\_id") \textbf{from} demographic inner join lab on demographic.hadm\_id = lab.hadm\_id \textbf{where} demographic."{admission\_type}" = "emergency" and lab."label" = "\textcolor{red}{troponin i}"  \\\hline

PtrGen+recover &\textbf{select} count ( distinct demographic."subject\_id" ) \textbf{from} demographic inner join procedures on demographic.hadm\_id = procedures.hadm\_id \textbf{where} demographic."gender" = "f" and \textcolor{red}{procedures."long\_title"} = "\textcolor{red}{spinal tap}"  &\textbf{select} count (distinct demographic."subject\_id") \textbf{from} demographic inner join lab on demographic.hadm\_id = lab.hadm\_id \textbf{where} demographic."{admission\_type}" = "emergency" and lab."label" = "\textcolor{red}{troponin i}"   \\\hline

TREQS & \textbf{select} count (distinct demographic."subject\_id") \textbf{from} demographic inner join procedures on demographic.hadm\_id = procedures.hadm\_id \textbf{where} demographic."gender" = "f" and procedures."short\_title" = "abdomen artery abdomen"  &\textbf{select} count (distinct demographic."subject\_id") \textbf{from} demographic inner join lab on demographic.hadm\_id = lab.hadm\_id \textbf{where} demographic."admission\_type" = "emergency" and lab."label" = "ferritin"  \\\hline

TREQS + recover &\textbf{select} count (distinct demographic."subject\_id") \textbf{from} demographic inner join procedures on demographic.hadm\_id = procedures.hadm\_id \textbf{where} demographic."gender" = "f" and procedures."short\_title" = "abdomen artery incision"  &\textbf{select} count (distinct demographic."subject\_id") \textbf{from} demographic inner join lab on demographic.hadm\_id = lab.hadm\_id \textbf{where} demographic."admission\_type" = "emergency" and lab."label" = "ferritin" \\\hline
\end{tabular}
} 
\label{tab:outsamples2} 
\end{table*}

\begin{table*}[!tp]
\vspace{-1mm}
\centering
\caption{Visualization of the accumulated attention on conditions that are used in the proposed TREQS approach on NL questions. Different conditions are labeled with different colors. An intense shade on a word indicates a higher attention weight.}
\vspace{-3mm}
\renewcommand\arraystretch{0.8}
\setlength\tabcolsep{1pt}
\resizebox{\linewidth}{!}{
\begin{tabular}{c}
\begin{minipage}{\textwidth}
\includegraphics[width=1.0\linewidth, height=72mm]{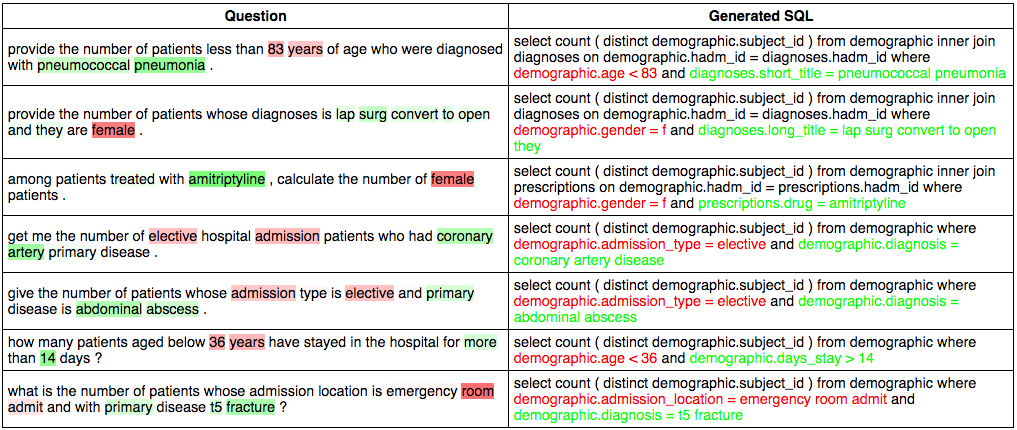}
\end{minipage}
\end{tabular}
} 
\vspace{-4mm}
\label{tab:vis} 
\end{table*}

It is easy to observe from Tables~\ref{tab:results1} to~\ref{tab:results-3} that the performance of condition value dominates the overall SQL generation performance. Seq2Seq is not able to capture the correct condition values due to its limitation in handling the OOV words. PtrGen performs slightly better since it is able to copy OOV words directly from the input questions, however, it still cannot capture the condition values as accurately as our proposed TREQS model. We believe that this is due to the fact that we consider temporal attention on questions, dynamic attention on SQL and the controlled generation and copying techniques in the proposed model. We can also observe that the proposed recover technique on the condition values can also improve the model performance significantly on both template questions and NL questions. As shown in Table~\ref{tab:results-noise}, the condition values can also be recovered effectively even if only partial condition value information is provided in the input questions. More analysis about the recover technique will be provided.

\vspace{-1.5mm}
\subsubsection{Analysis of the Generated SQL Query}
In addition to the quantitative evaluations, we have also conducted an extensive set of qualitative case studies to compare the SQL queries produced by various models. Two examples on NL questions are provided in Table~\ref{tab:outsamples2}.
For both examples, different comparison models have given correct answers for the template, table name, and columns.
Note that the Coarse2Fine model cannot handle questions on two tables. 
For example 1, M-SQLNET provides a wrong procedure short title ``\textit{parent infus nutrit sub}'' due to the mis-classification error. 
Seq2Seq and Ptr generate a partially correct procedure short title ``\textit{other abdomen}'' and ``\textit{spinal abdomen artery}'', respectively. In addition to their inability to obtain the correct condition values, these baseline methods do not even have the ability to correctly predict the condition column ``\textit{procedures.short\_title}'' in example 1. 
Similarly, in example 2, the generated SQL query by Seq2Seq model is not executable even if it correctly predicts the value ``\textit{ferritin}'' for the second condition, since it predicts an incorrect condition column ``\textit{lab.itemid}''. In this case, the recover technique can only recover the condition value ``\textit{51200}'' for ``\textit{lab.itemid}'' instead of keeping the condition value ``\textit{ferritin}'' that is correctly generated. 
These predicted results indicate that successfully recovering the condition values still requires the language generation models to produce correct condition columns and sufficiently relevant keywords.
Note that it is unable to recover the condition values for M-SQLNET since its predicted values are already in the look-up table.
Different from these baseline methods, our proposed TREQS model is able to generate totally correct SQL queries for both examples even without applying the recover technique. This shows the efficiency of our TREQS method in predicting the correct condition values without affecting the performance of other components in the SQL query.

\vspace{-2.5mm}
\subsubsection{Accumulated Attention Visualization}
Visualization of attention weights can help in interpreting the model and explaining experimental results by providing an intuitive view about the relationships between generated tokens and source context, i.e., input questions.
In Table~\ref{tab:vis}, we show seven natural language examples with reasoning questions and SQL queries that are generated using the proposed TREQS method.
The goal here is to investigate if TREQS is able to successfully detect important keywords in a question when generating conditions in its corresponding SQL query. 
Therefore, we choose to visualize the accumulated attention weights instead of the weights for each of the generated tokens.
For example, for the question ``\textit{get me the number of elective hospital admission patients who had coronary artery primary disease}'', the model mainly focuses on ``\textit{elective}'' and ``\textit{admission}'' when generating condition ``\textit{demographic.admission\_type = elective}'', and on ``\textit{coronary artery}'' when generating ``\textit{demographic.diagnosis = coronary artery disease}''. In this example, the condition values are mainly obtained by directly copying from the input question since they are explicitly included. On the other hand, the condition value ``\textit{f}'' in the SQL query for question ``\textit{among patients treated with amitriptyline, calculate the number of female patients}'' is mainly obtained through the controlled generation and copying technique since ``\textit{f}'' is not explicitly provided in the input question. 
Similarly, TREQS model is able to capture relevant keywords for each condition in other examples. 

\vspace{-1mm}
\section{Conclusion}
\label{sec6}
Large amounts of EMR data are collected and stored in relational databases at many clinical centers. Effective usage of the EMR data, such as retrieving patient information, can assist doctors in making future clinical decisions. Recently, the Question-to-SQL generation methods have received a great deal of attention due to their ability to predict SQL query for a given question about a database. Such an automated query generation from a natural language question is a challenging problem in the healthcare domain. In this paper, based on the publicly available MIMIC~III dataset, a Question-SQL pair dataset (MIMICSQL) is first created specifically for the Question-to-SQL generation task in healthcare. We further proposed a TRanslate-Edit Model for Question-to-SQL~(TREQS) generation task on MIMICSQL by first generating the targeted SQL directly and then editing with both attentive-copying mechanism and a recover technique. The proposed model is able to handle the unique challenges in healthcare and is robust to randomly asked questions. Both the qualitative and quantitative results demonstrate the effectiveness of our proposed method.
\vspace{-1mm}

\begin{acks}
This work was supported in part by the US National Science Foundation grants IIS-1619028, IIS-1707498 and IIS-1838730.
\end{acks}

\newpage
\bibliographystyle{ACM-Reference-Format}
\balance 
\bibliography{7-reference}

\end{document}